\documentclass[a4paper,fleqn]{cas-sc}
\usepackage{caption,setspace}

\usepackage{graphicx}
\usepackage{caption}
\usepackage{mwe}
\RequirePackage{expl3}
\usepackage{cite}
\usepackage{amsmath,amssymb,amsfonts}
\usepackage{algorithmic}
\usepackage{graphicx}
\usepackage{textcomp}
\usepackage{slashbox}
\usepackage{xcolor}
\usepackage{booktabs}
\usepackage{natbib}
\usepackage[labelfont=bf]{caption}
\usepackage{url}
\usepackage{multirow}
\usepackage{boldline}
\usepackage{subcaption}
\usepackage{caption,booktabs}
\usepackage{makecell}
\usepackage{soul}

\begin{document}
\let\WriteBookmarks\relax
\def\floatpagepagefraction{1}
\def\textpagefraction{.001}

\shorttitle{Document-Level Sentiment Analysis of Low-Resource Language}


\title [mode = title]{Document-Level Sentiment Analysis of Urdu Text Using Deep Learning Techniques}                      
\tnotemark[1]
\tnotetext[1]{This work is not sponsored/supported by any organization.}


%
\author[1]{A Irum}[type=editor,
                        ]



\ead{airum.msit18seecs@seecs.edu.pk}


\credit{Methodology, Data curation, Writing - Original draft preparation, Visualization}

\affiliation[1]{organization={School of Electrical Engineering and Computer Science (SEECS), National University of Sciences and Technology (NUST)},
    addressline={H-12}, 
    city={Islamabad},
    country={Pakistan}}
\cormark[1]
\author[1]{M Ali Tahir}
                    [
                        orcid=0000-0002-2335-2776
                        ]

\fnmark[1]

\ead{ali.tahir@seecs.edu.pk}
\cormark[1]
\credit{Formal analysis, Supervision, Writing – review \& editing, Project administration, Validation}
\author[1]{S Latif}

\fnmark[1]
\ead{seemab.latif@seecs.edu.pk}

\credit{Review \& editing, Validation}




\begin{abstract}
Document level Urdu Sentiment Analysis (SA) is a challenging Natural Language Processing (NLP) task as it deals with large documents in a resource-poor language. In large documents, there are ample amounts of words that exhibit different viewpoints. Deep learning (DL) models comprise of complex neural network architectures that have the ability to learn diverse features of the data to classify various sentiments. Besides audio, image and video classification; DL algorithms are now extensively used in text-based classification problems. To explore the powerful DL techniques for Urdu SA, we have applied five different DL architectures namely, Bidirectional Long Short Term Memory (BiLSTM), Convolutional Neural Network (CNN), Convolutional Neural Network with Bidirectional Long Short Term Memory (CNN-BiLSTM), Bidirectional Encoder Representation from Transformer (BERT). In this paper, we have proposed a DL hybrid model that integrates BiLSTM with Single Layer Multi Filter Convolutional Neural Network (BiLSTM-SLMFCNN). The proposed and baseline techniques are applied on Urdu Customer Support data set and IMDB Urdu movie review data set by using pretrained Urdu word embeddings that are suitable for (SA) at the document level. Results of these techniques are evaluated and our proposed model outperforms all other DL techniques for Urdu SA. BiLSTM-SLMFCNN outperformed the baseline DL models and achieved 83{\%}, 79{\%}, 83{\%} and 94{\%} accuracy on small, medium and large sized IMDB Urdu movie review data set and Urdu Customer Support data set respectively.
\end{abstract}




\maketitle

\section{Introduction}
With the tremendous increase in the availability of textual data in recent years, wide interest of the research community has been seen in different new and relatively more demanding areas in the field of Natural Language Processing (NLP). One such active research area of NLP is Sentiment Analysis (SA). SA tends to analyze a wide range of opinions regarding various subjects that are expressed by humans on different platforms. The objective of SA is to identify the sentiment and assign it a label in accordance with the particular subject \cite{liu2012sentiment}. SA has gained massive attention recently as sentiment is considered the key component in maintaining the reputation of organizations associated directly with public. 

English language has gained much attention for global-level research since the birth of NLP. In the last two decades, wide-scale research regarding each aspect of NLP has mostly been conducted in English language as it is computationally inexpensive to conduct experimentation in a resource-rich language. As compared to resource-rich languages, morphologically rich yet resource-scarce languages like Urdu, Arabic, Turkish, Hindi and Persian, etc. have been least utilized by researchers for NLP tasks as it is hard to cater word-level complex morphological structure of such languages ~\cite{abdul2010automatic}. The unique characters, complex morphology, and scarcity of linguistic resources of Urdu language have played a vital role in limited research regarding Urdu SA \cite{daud2017urdu}. 

 Most of the traditional work done in Urdu SA domain is based on  Machine Learning (ML) models. ML models do not achieve the best results on large data sets and tend to learn only directed text features. The significant challenge regarding ML models is the selection of efficient features from the high-dimensional feature space [4]. Among the numerous feature selection methods available, there is no single method that works well with all types of ML classifiers \cite{zia2015comparative}. 

The performance of Deep Learning (DL) models have improved on a large scale with a considerable amount of data compared to traditional ML models \cite{mathew2021deep}. Moreover, DL models have the ability to learn complex features directly from high-dimensional feature space and they are faster than ML models because in DL models, parallel processing of data is done by using Graphic Processing Units (GPUs) \cite{khan2019deep}. Unfortunately, the potential of DL models has not yet been explored in the field of Urdu SA. 

In the last few years, DL models have proved to show the best results, particularly in the domain of sentiment analysis. Among these models, Convolutional Neural Network (CNN) and Recurrent Neural Network (RNN) have been widely implemented as CNN responds very well to the problem of dimensionality reduction and a type of RNN, the BiLSTM network, handles sequential data with success \cite{goularas2019evaluation}. In this study, we present a hybrid Bidirectional Long Short Term Memory with Single Layer Multifilter Convolutional Neural Network  (BiLSTM-SLMFCNN) that combines BiLSTM and multifilter CNN architecture to classify sentiments in Urdu documents. The major advantage of integrating BiLSTM and CNN is to utilize the potential of BiLSTM that fully understands the meanings of individual words in the context from both directions and the capability of the multifilter CNN architecture to extract local features in high-dimensional feature space. Moreover, CNN ignores the contextual meaning of words in text-based classification, hence, BiLSTM is integrated to improve the accuracy of the hybrid model for Urdu sentiment classification. The proposed scheme takes advantage of sequential text processing and extraction of global features simultaneously. The novelty of our proposed technique lies in the use of SLMFCNN layer as it contains multiple filters of different sizes which are applied to extract variable length features, i.e. n-grams from text documents.

\subsection{URDU AND ITS ATTRIBUTES:}
Urdu is the national language of Pakistan and it is world-wide notable language with 300 million speakers all across the globe \cite{riaz2012comparison}. It is 21\textsuperscript{st} most spoken language in the world. It has a character set of 38 characters shown in Figure 1. Along with the complex morphological structure, it has the following distinctive features that make it a challenging language for computational tasks \cite{syed2011sentiment}:
    
\begin{itemize}
    \item It is a mixture of different languages and has a lot of loan words from languages like Turkish, Arabic, Persian, Sanskrit, and even English. It has significant influence from other languages.
    
    \item It has morphological variations, i.e., many words have a common root word. 
    
    \item Urdu has Nastalique writing style as shown in Figure 1. Nastalique writing style is complex in its nature.

    \begin{figure}[!ht]
    \centering
    \includegraphics[width=0.5\linewidth, height=0.5\linewidth]{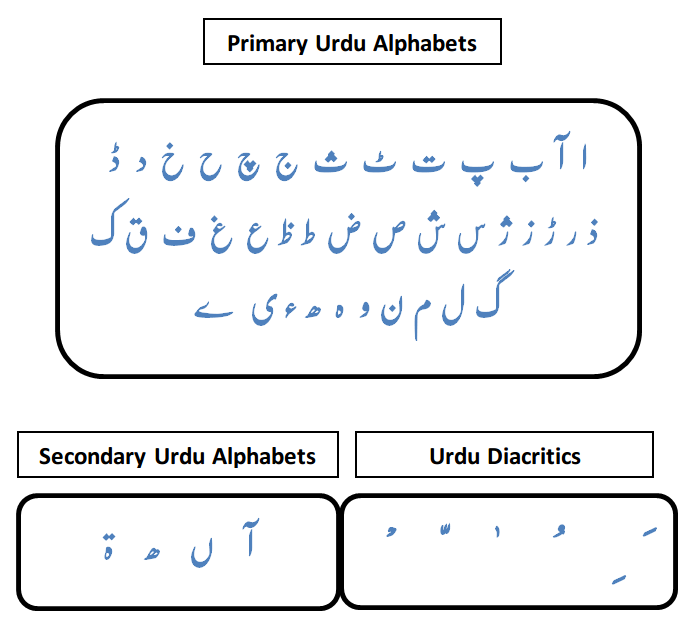}
    	\caption{\textbf{Alphabets and Diacritics of Urdu Language in Nastalique Font Style}}
    	\label{fig:Figure 1:}
    \end{figure}
    
    \item Urdu script is context-sensitive. Context-sensitivity outcomes in problems related to word segmentation as the space between complete words does not always indicate word boundaries. Such words can be misidentified. 
    \item There is no concept of capitalization in Urdu language. Hence, due to this feature, beginning of the sentence and proper nouns cannot be easily identified \cite{akhter2022exploring}.
    
    \item In Urdu, case markers are considered as parts of speech. Case markers are lexically independent units, hence they determine the sentence structure, resulting into grammar related ambiguities.
    
    \item Use of diacritics also change the meaning of words having same spelling or pronunciation \cite{akhter2020document}.
\end{itemize}

\subsection{CATEGORIZATION OF SENTIMENT ANALYSIS LEVELS:}
SA is categorized into three contrasting levels, namely word-level, sentence-level and document-level \cite{rhanoui2019cnn}. At word-level, the polarity of word is determined i.e. whether the word is positive, neutral or negative.
At sentence-level SA, the polarity of whole sentence is calculated in order to examine the sentiment score of sentence. A sentence is a sequence of words that define the opinion of a person on a particular subject. It is usually done for data extracted from social networking websites. Furthermore, at document level, SA aims to classify the polarity of complete document by taking into consideration an increase of words and noise features that tend to distort the learning process resulting in difficulty in sentiment prediction of the document.

\subsection{CHALLENGES REGARDING DOCUMENT-LEVEL URDU SENTIMENT ANALYSIS:}
Besides the complex morphological structure and unique features of Urdu language, one of the key challenges met by researchers in applying Urdu SA is the availability of an annotated Urdu corpus. Scarcity of large, free of cost and publicly available standard corpus makes it difficult for researchers to expand the scope of their research in the field of Urdu SA. This obstacle has created hindrance in the application of text classification of Urdu text, particularly at document-level. Moreover, DL approaches are still unexplored in the domain of document-level Urdu SA.  
To fill these gaps, the main contributions of this study are: 
\begin{itemize}
    \item A hybrid model BiLSTM-SLMFCNN is developed for document-level Urdu SA.
    \item The performance of proposed and baseline DL models is analyzed on variable sized data sets.
    \item The impact of data size on the performance of proposed and baseline DL models is evaluated.
    \item Comparative analysis regarding to the performance of BiLSTM-SLMFCNN is done with well-known DL classifiers.
\end{itemize} 

The organization of paper is as follows: Section II presents literature review. Section III briefly explains the baseline DL architectures. In section IV, we have presented our proposed model. Section V is related to detailed description of experimental procedure including comparison of two data sets, preprocessing steps, experimental setup and evaluation metrics. In section VI, results and discussion are included. Section VII presents conclusion and future work.\\

\begin{figure*}[!ht]
\centering
\includegraphics[width=0.7\linewidth, height=0.4\linewidth]{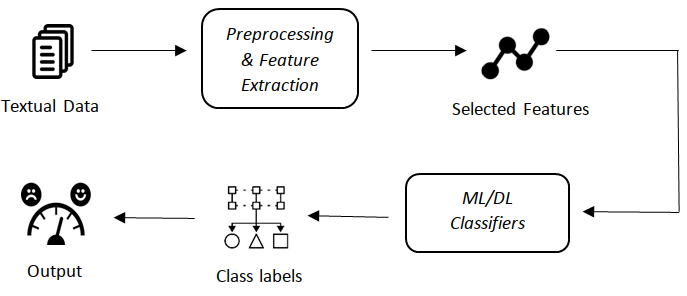}
	\caption{\textbf{Sentiment Analysis Framework}}
	\label{fig:Figure 2}
\end{figure*}

\section{LITERATURE REVIEW}
\label{sec:Literature Review}

Classification of Urdu text by making use of lexical resources, feature selection methodologies and ML classifiers has been in progress since the last decade \cite{ahmed2016framework}. As different languages have contrasting dialect, morphological and lexical structure, the methods adopted for SA of resource-rich languages cannot be utilized for Urdu language \cite{syed2010lexicon}. Almost a decade ago, researchers started to focus on exploitation of the concept of Urdu SA. From Bag-of-Words (BoW) model to supervised ML algorithms, Urdu SA has evolved much over the years. However, there is still a research gap when it comes to Urdu SA by applying mainstream DL methodologies.

\cite{khattak2021survey} conducted an elaborated survey on Urdu SA. According to their research, Urdu has faced lack of interest from researchers' side because of issues such as segmentation, morphological dissimilarities, lexical inconsistencies and scarcity of acknowledged resources. Urdu SA has been burdened with a lot of shortcomings, for instance, lack of a gold-standard corpus, deficiency of sentiment lexicons, ineffective handling of negations and modifiers, complications in management of domain-specific words, slang and emoticons recognition etc.

\cite{ali2009urdu} evaluated the performance of two ML classifiers, Naive Bayes (NB) and Support Vector Machine (SVM) for the task of classifying Urdu text documents. They manually prepared a new data set by scrapping several Urdu news websites and manually annotated them into six classes i.e. sports, news, culture, finance, personal and consumer information. Their experimental results showed that SVM outperformed NB with a significant margin. Their results also revealed that the process of stemming reduced the general performance of classification process.

By making use of discourse (sub-sentence) level information, \cite{awais2019role} proposed to improve the performance of Bag of Words (BoW) model and various ML models for Urdu SA. Their proposed methodology works in a way that firstly, discourse data is extracted, then new set of rules are defined for rule-based classifier which tends to ultimately improve the performance of BoW model and lastly, new set of features for SVM were introduced. The results of their proposed methodology suggest that due to the use of context level information, ML classifier outperformed rule-based classifier by 5.16\text{\%}.However, classification accuracy is dependent on the number of sub-opinions in the sentence. For more than two sub-opinions, the proposed technique will show limited accuracy. 

\cite{hassan2018opinion} have carried out Urdu SA using the sub-opinion level details. They proposed a systematic model named SEGMODEL to classify sub-sentence level information into positive and negative classes. They have also explored the capabilities of BoW model as it works well for analyzing polarity of a simple sentence but it tends to fail on the SA of complex sentences. The SEGMODEL proves the hypothesis that overall polarity of sentence depends on second sub-sentence. The results also showed that segmentation increases the overall performance of SA framework. The proposed model enhanced the accuracy by 24.75\text{\%}. However, the proposed model is not compared with well-known classifiers.

Abundant amount of research has been carried out by making use of sentiment lexicons. The sentiment lexicons refer to a repository comprising of sentiment carrier terms with sentiment category and polarity \cite{asghar2017cogemo}. Both manual \cite{melville2009sentiment} and automatic mechanisms of generation \cite{bloom2010automated} of lexicons has been in focus of research community. This approach lags behind machine learning approaches because of lack of sentiment bearing lexicons that can deal with dialect issues \cite{nassr2019comparative}. Lexicon based approaches compare each term in the corpus with lexicon entries. As an output, orientation score is attached to each term on the basis of which polarity of whole sentence is calculated \cite{syed2011adjectival}. 

\cite{amjad2017exploring} devised an algorithm for application of SA for Urdu news gathered from Twitter. They gathered 26000 tweets from account handles of Pakistan’s leading newspaper groups. They tokenized the data to create a lexicon from data and remove duplicate words. The list of unique words was then refined to serve as lexicon. POS tagging was done to figure out adjectives, adverbs and nouns. These words were referred as effective words while rest of the words was called ineffective. Labeling of training data was done by human annotators and final label was selected by majority voting. Their proposed algorithm works in a manner that sentiment polarity score calculation is followed by segmentation and then filtration. Polarity score was assigned to effective words after the process of segmentation and on its basis, sentiment of news tweet was checked. Their proposed algorithm produced 77\text{\%} accurate results. The proposed scheme ignores preprocessing steps and is generic in nature.

Urdu tweets have been used as data set to test the accuracy of Decision Tree (C45) on Urdu text. They have collected 600 tweets from Twitter from which 500 were separated as train data \cite{bibi2019sentiment}. The training data was labeled by human annotators. Preprocessing techniques including text cleaning and stopwords removal have been applied. After preprocessing, features vector was developed by making use of positive and negative words, POS tags and use of negation in tweets. Adjective is a part of speech tag that helps determine the polarity of text. So, adjective with positive sense was scored as +1 while an adjective with negative sense was scored as -1. The evaluated results of proposed scheme showed 90\text{\%} accuracy for SA of Urdu tweets. The ML model achieved good accuracy, however, comparative analysis of different ML models have not been done for Urdu tweets.

For the purpose of figuring out best ML classifier in the domain of Urdu SA, \cite{mukhtar2018identification} compared five classifiers namely Naive Bayes (NB), Lib SVM, KNN (IBK), Decision Tree (J48) and PART through WEKA. Three standard measures of evaluation i.e. Kappa measure, root mean squared error and McNemar’s test were employed to verify the results. They have applied sentence-level SA on 151 Urdu blog posts categorized into fourteen diverse genres having three standard sentiments i.e. positive, neutral and negative. Stopwords were manually removed for getting accurate results. Classification was done through WEKA with help of attribute relation file format. 39 attributes were extracted to be used as features for sentences. After getting results of all classifiers, their performance was evaluated. According to the results, KNN (IBK) gave best results hence it was declared best classifier by authors for sentence-level Urdu SA.

\begin{figure}[!ht]
\centering
\includegraphics[width=0.6\linewidth, height=0.6\linewidth]{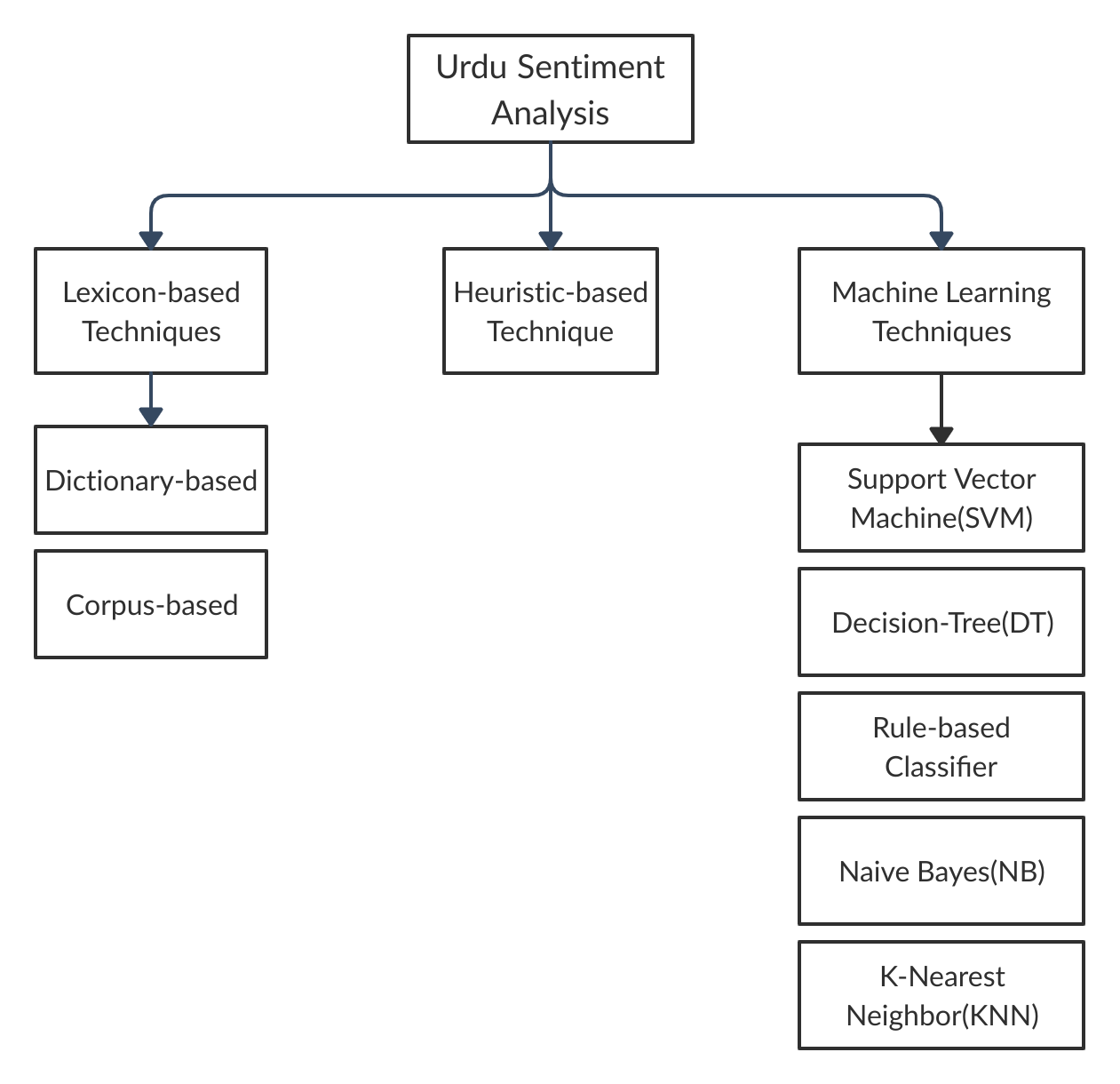}
	\caption{\textbf{Taxonomy of Techniques Applied for Urdu Sentiment Analysis}}
	\label{fig:Figure 3}
\end{figure}

While analyzing sentiments for Urdu text, intensifiers hold a significant place. \cite{mukhtar2017effective} proposed a methodology in which intensifiers have been used to enhance the accuracy of classification process. If the intensifier is surrounded by either a positive or a negative word, they have carefully specified rules for assigning polarities to intensifiers particularly while developing Urdu Sentiment Analyzer. 6025 sentences have been collected from 151 blog posts. A lexicon was created to assign polarity to positive and negative terms from data. By using their own lexical resource along with rule-based mechanism for Urdu sentiment classification, their proposed methodology achieved 83\text{\%} accuracy. They also compared accuracy of supervised ML models with their sentiment analyzer which showed that Urdu Sentiment analyzer is more accurate than supervised machine learning classifiers.  Lexicon based approach is easy to implement but lexicon creation itself is a tedious task as compared to implementation of DL models.

\cite{ali2016salience} performed salience analysis using heuristic based approach in order to find out degree of polarity of Urdu political news. They have used following corpora for experimentation i.e. raw corpus; POS tagged corpus and manually tagged corpus. Word lists having polarity were also maintained for carrying out salience analysis. After tokenization, POS tagging was applied for finding out saliences i.e. entities carrying positive, negative or neutral sentiment. Using this approach, they figured out significant entities by making use of parameters like proper nouns, repetitive words. Calculating overall polarity gives the final sentiment for a particular news entity. Their results showed that positive saliences are lesser than negative saliences. Their proposed methodology achieved 84.5\text{\%} accuracy.

In a nutshell,  researchers have applied lexicon-based, heuristic-based and Machine Learning based techniques. Deep Learning techniques for Urdu SA is still in the phase of inception. A considerable amount of work has been done using Machine Learning and Deep Learning approaches for Roman Urdu \cite{ghulam2019deep} and other resource-rich languages. For Sentiment Analysis of native Urdu Language, the Deep Learning area is still to be explored. The application of our proposed architecture and baseline Deep Learning models tend to address the existing gap as mentioned in the literature by adopting a hybrid Deep Learning approach that involves CNN layer with multi-size filters  followed by BiLSTM layer. The proposed technique provides higher classification accuracy on different sized data sets.

\begin{table*}[H]%
\centering
\setlength{\tabcolsep}{2.5pt}

\begin{tabular*}{\textwidth}
{@{\extracolsep\fill}lccccccccr@{\extracolsep\fill}}
\hline
\textbf{S.}\textbf{No} & \textbf{Study} & \textbf{Approach} & \textbf{Level} & \textbf{Data Set} & \textbf{Accuracy} \\
\hline
\addlinespace[0.3cm]
1 & Ali et al. & Machine Learning & Document & News & 5.16\text{\%}\\
\addlinespace[0.3cm]
2 & Awais et al. & BoW and  Machine Learning & Sub-sentence & Product Reviews & Improved by  5.16\text{\%}\\
\addlinespace[0.3cm]
3 & Hasan et al. & BoW and SEGMODEL & Sub-sentence & Product Reviews & Improved by 24.75\text{\%}\\
\addlinespace[0.3cm]
4 & Amjad et al. & Sentiment Annotated Lexicon
 & Sentence & Tweets & 77{\%} \\
\addlinespace[0.3cm]
5 & Bibi et al. & Machine Learning & Sentence & Tweets & 90\text{\%} \\
\addlinespace[0.3cm] 
6 & Mukhtar et al. & Machine Learning & Sentence & Blogs & 67.0185\text{\%}\\
\addlinespace[0.3cm]
7 & Mukhtar et al. & Sentiment annotated Lexicon & Sentence & Blogs & 83\text{\%} \\
\addlinespace[0.3cm]
8 & Abbas et al. & Heuristic Approach & Sentence & News & 84.5\text{\%}\\

\addlinespace[0.3cm]
\hline
\end{tabular*}
\centering
\caption{\textbf{Comparative Analysis of Existing Work }}
\end{table*}  

\section{DEEP LEARNING MODELS-A BACKGROUND}

DL is the advanced area of ML wherein Artificial Neural Networks (ANN), which are inspired from human neural system, are used to learn features of huge observational data in order to predict about unseen test data. Deep learning models have been extensively deployed by researchers for SA and Opinion Extraction on resource-rich languages. The results of these models have strengthened the trust of researchers on Artificial Intelligence (AI). In this section, DL models RNN, CNN and BERT are explained in detail.

\subsection{RECURRENT NEURAL NETWORKS:}
Recurrent Neural Networks (RNN) is a sequence learning model in which inputs are interconnected. It represents the generalized form of feed forward NN in which nodes between model’s hidden layers are connected and sequence features are learnt dynamically. 
RNN learns context of the sentence while training. Semantic information between words can be transferred by RNN but it is not capable of capturing long distance semantic link between different words. During the process of training the model, the gradient gradually gets decreased until it vanishes completely. Consequently, length of the sequential data becomes limited.

\subsubsection{LONG SHORT-TERM MEMORY}
In order to cater the issue of RNN, Schmidhuber and Hochreiter proposed Long Short-Term Memory (LSTM) model that tends to learn long-term dependencies between different words by making use of three gates namely input gate, forget gate and output gate \cite{hochreiter1997long}. The input gate accepts or blocks the sequential input, the forget gate enables or disables a neuron based on the weights being learned by the model. The output gate determines output value of the LSTM’s unis.

\subsubsection{BIDIRECTIONAL LSTM:}
The traditional RNN model and LSTM can only propagate information in forward direction. This tendency allows these models to depend upon information processed before certain time. To cater this problem, Bidirectional LSTM is used \cite{schuster1997bidirectional}. It has been proven very helpful in situations where context of input is required. It tends to process data in two directions i.e. forward to backward and backward to forward, as it uses two hidden layers. By making use of two directions of time, input data from both past and future of the current time period is utilized to predict better results.

\subsection{CONVOLUTIONAL NEURAL NETWORK:}
The core concept of Convolutional Neural Networks (CNN) is local features extraction, weight distribution and down sampling. It captures local correlation between neighboring words by focusing on local connectivity patterns amongst adjacent layer neutrons. It can easily learn text features effectively from massive amount of text using 1-dimensional or 2-dimensional (word order) structure using convolutional layers. The convolution layer makes use of convolving filters of variable sizes in order to extract high-level textual features. Then, maxpooling layer is added to extract global features. Both convolutional and max pooling layers make it possible for the model to learn to figure out local indicators while staying indifferent to their position in sentence/document \cite{rakhlin2016convolutional}. One major challenge regarding CNN is to figure out suitable number
of filters and filter size. Large-sized filters effects the training process
while fewer filter results in incorrect results.

\subsection{BIDIRECTIONAL ENCODER REPRESENTATION FROM TRANSFORMERS (BERT):}

Bidirectional Encoder Representation from Transformers (BERT) is a deep bidirectional model for general purpose language understanding that learns information from both left to right and right to left directions. It has largely been utilized in the domain of sentiment analysis \cite{gao2019target}. BERT uses transformer model that applies attention mechanism and encoder-decoder model on NLP tasks. The attention procedure helps the model to concentrate on related parts of given input sequence \cite{vaswani2017attention}. Both encoder and decoder separately have some underlying understanding of language and because of this understanding; transformer architecture can be used to build systems that perfectly understand language. 

\section{PROPOSED MODEL:}
\label{sec:Proposed Model}

\begin{figure*}[!ht]
\centering
\includegraphics[width=1.0\linewidth, height=0.5\linewidth]{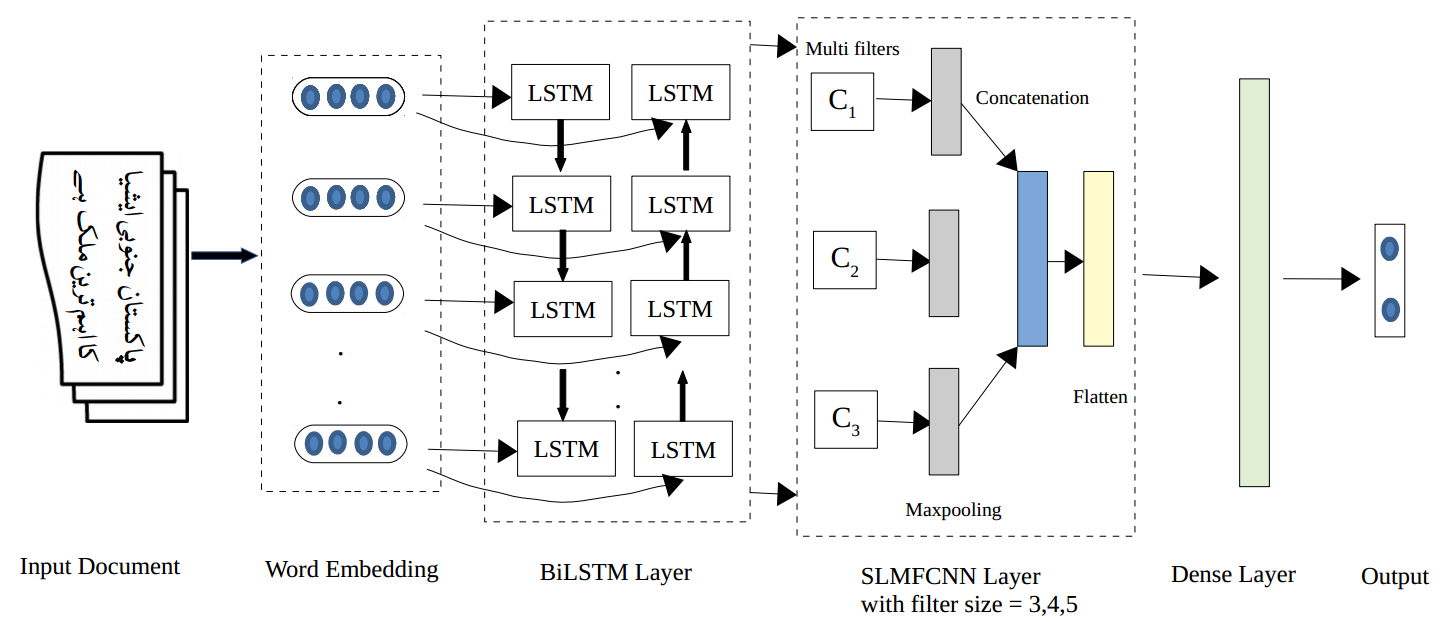}
	\caption{\textbf{BiLSTM with Single Layer Multi Filter Convolutional Neural Network (BiLSM-SLMFCNN)}}
	\label{fig:Figure 4}
\end{figure*}

The proposed model combines two widely used NN models namely, BiLSTM and CNN. Figure 4 illustrates the proposed BiLSTM-SLMFCNN model. This hybrid model is combined in order to test the adaptation of BiLSTM with CNN, as both of them are renowned for their use in sentiment analysis. BiLSTM and CNN have diverse purposes in SA and they are considered to be the mainstream models for classifications tasks. BiLSTM classifies the emotions in text by using the semantics of textual sequence and it maintains the chronological order amid words in the document, hence it is capable to ignore unwanted words by using its delete gate. CNN extracts the latent semantic evidence of text by convolving the embedding text. It can extract as numerous textual features from the text as possible. Based on these facts, our research combines the above mentioned two models and proposes a hybrid DL model based on BiLSTM and CNN. 

As compared to the traditional BiLSTM-CNN model, the BiLSTM-SLMFCNN model focuses on the association between textual features. In this research, we aim to find out that by combining the textual features extracted by BiLSTM and CNN algorithms, we can efficiently solve the issues related to Urdu sentiment classification.  

The purpose behind combining these two neural network models is to generate a hybrid model that takes benefit of the strengths of both CNN and BiLSTM, so that it filters the information using BiLSTM, and uses them as CNN input. Thus, we present a model BiLSTM-SLMFCNN that meets this goal.

A single input channel is used with predefined Urdu word embedding such that the word embedding vectors are used as BiLSTM input. Then, three filters of sizes 3, 4 and 5 are applied for 100 times each to obtain multiple feature maps of each variable-sized filter. After applying each filter, maxpooling layer is applied on each feature map to reduce and update the data size. The output of maxpooling layers is concatenated and passed to flatten layer. It converts the pooled features map into a single column and passes it to the dense layer. The dense layer uses the softmax activation function to classify sentiments. Dense layer is added to balance and optimize the input dimensions so that the proposed model can extract the features in the text in a better way. The hyperparameters of BiLSTM-SLMFCNN like dropout rate, batch size, activation function and learning rate were fine-tuned using grid-search technique. Optimized parameters help to minimize the loss and improve the overall performance of model.
Besides our proposed model; BiLSTM, CNN, CNN-BiLSTM and BERT have also been applied on VTC and IMDB movie reviews data set.

\section{EXPERIMENTAL STUDY:}
\label{sec:EXPRIMENTAL STUDY:}
\subsection{Data Sets:}

The data set used in this research has been provided by a Pakistani Vehicle Tracking Company (VTC). The data set is related to company’s customer support center calls where customers talk to company’s customer service representative for their queries and complaints. This speech data is anonymized and transcribed into textual data. It is an annotated, small data set with 405 calls categorized as satisfied and 100 calls categorized as unsatisfied.
\begin{table*}
\centering
\begin{tabular}{ccccc}
\toprule
\addlinespace[0.3cm]
Properties & VTC & IMDB Small & IMDB Medium & IMDB Large \\
\midrule
\addlinespace[0.3cm]
Size & Small & Small & Medium & Large\\ 
\addlinespace[0.3cm]
No of Docs & 505 & 600 & 3000 & 10000\\ 
\addlinespace[0.3cm]
Class Ratio & 
\multicolumn{1}{@{}c@{}}{
   \begin{tabular}{c} 
         405 Satisfied  \\
         100 Unsatisfied \\
   \end{tabular}
} & 
\multicolumn{1}{@{}c@{}}{
   \begin{tabular}{c}
         306 Positive  \\
         294 Negative  \\
   \end{tabular}
} & 
\multicolumn{1}{@{}c@{}}{
   \begin{tabular}{c}
         1543 Positive  \\
         1457 Negative \\
   \end{tabular}
} & 
\multicolumn{1}{@{}c@{}}{
   \begin{tabular}{c}
         5033 Positive  \\
         4967 Negative \\
   \end{tabular}
} \\ 
\addlinespace[0.3cm]
Level of Imbalance & High & Low & Low & Low\\
\bottomrule
\end{tabular}
\caption{\textbf{Comparative Analysis of Data Sets}}
\end{table*}
Besides VTC’s data set, IMDB movie review data set\footnote{https://www.kaggle.com/akkefa/imdb-dataset-of-50k-movie-translated-urdu-reviews} in Urdu language has also been used in this research for comparison and evaluation. It is a translated data set of original IMDB English movie review data set and it has been translated using Google Translator. The reason for using this  is that it has high polarity for both the classes. This data set has 50,000 movie reviews that are equally divided into two classes, positive and negative. Figure 10 and 11 show a portion of both data sets used for Urdu SA. For the sake of comparison with VTC’s data performance on the basis of data size, we have made three different groups of this data, namely IMDB Small, IMDB Medium and IMDB Large. Details of both data sets are given in Table 2.

\begin{figure*}[!ht]
\centering
\includegraphics[width=0.9\linewidth, height=0.18\linewidth]{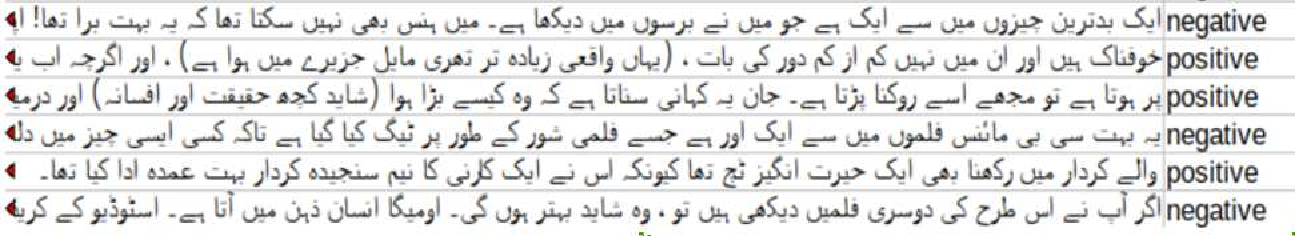}
	\caption{\textbf{Sample of IMDB Movies Review Data Set}}
	\label{fig:Figure 5}
\end{figure*}

\subsection{Preprocessing:}
Before using the text for any classification task, it is preprocessed so that clean, normalized and structured data is used for getting accurate results. Preprocessing helps to maintain data in a form that it has no redundancy and noise. The process is widely adapted by researchers to get cleaned data for better interpretation of applied model. In the implementation of our proposed model, we have applied text preprocessing to all the data sets for noise removal. First of all, tokenization is done in which each document’s word is considered as a separate token. Then Urdu stop words are removed in order to make the data less prone to redundancy. Then, further preprocessing is done by removing any English alphabets, alphanumeric characters, URL’s, whitespaces and punctuation marks. The preprocessed data is then fed into neural network so that model’s performance can be evaluated. Common preprocessing steps are shown in Figure 9.

\begin{figure*}[!ht]
\centering
\includegraphics[width=0.9\linewidth, height=0.3\linewidth]{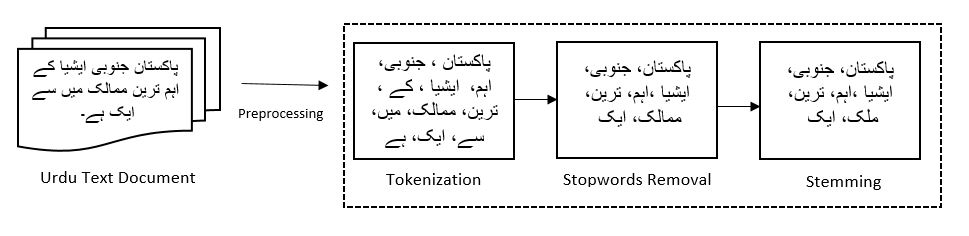}
	\caption{\textbf{Common Text Preprocessing Steps }}
	\label{fig:Figure 6}
\end{figure*}

\subsection{Experimental Setup:}
After the process of data acquisition and transcription, data preprocessing is done. Urdu Preprocessing is itself a tedious task but it has been made easy by Urduhack\footnote{https://pypi.org/project/urduhack/}  library. For parameters setting, we have implemented grid search technique and on basis of careful implementation, we have applied 3-fold cross validation approach to find out optimized parameters for our proposed model and other baseline models namely BiLSTM, CNN, CNN-BiLSTM (Input layer, Urdu Word Embedding layer, convolutional layer with filter size 5 and 100 feature maps, BiLSTM layer, Dense Layer) and BERT. The hierarchy of layers is as follows:

\subsubsection{\textbf{Word Embedding layer: }}
The pre-processed data set gives a meaningful and unique sequence of words and every word is assigned a unique ID. Word embedding layer assigns random weights to words and it eventually learns the embedding to embed all the words in the training data set. This layer is mostly used to learn word embeddings that can be saved for further use in any other model. In this study, we have used pre-trained Urdu word embeddings \cite{haider2018urdu} trained on skipgram model in the embedding layer. 

\subsubsection{\textbf{BiLSTM Layer: }}
BiLSTM layer takes the input of word vectors from embedding layer. It tends to keep the sequential order between the text data. It precisely detects the links between previous inputs and outputs. 

\subsubsection{\textbf{Convolutional Layer: }}
A convolutional layer works as a feature-extracting module that aims to explore the combination between different sentences of the document by using multiple filters of size \textit{t}. In this layer, multisize filters act as n-gram detectors where each filter looks for the particular class of n-grams and then assigns high scores to them. Those detected n-grams which have highest score then pass the max pooling stage. This layer uses multi filters of sizes 3, 4 and 5 which are applied 100 times each.
After application of each filter, max pooling operation is implemented to reduce and update the data size. The outcome of all max pooling layers are then concatenated to serve as input to Dense layer.

\subsection{Methodology: }
In the proposed model BiLSTM-SLMFCNN, we have used three filter sizes of (3, 4 and 5) along with 100 feature maps. The number of CNN filters
and the filter size affects the network performance and depends on the complexity of the problem in hand 
and the dataset. A large filter size usually slows down the training process while a small filter size decreases the performance.

The proposed model comprises of two sets of vectors, one is an input layer of the network and second one is the word embedding layer. The main notion is to capture the hidden semantic information in every word of each document. For this purpose, we have used pretrained neural word embedding proposed by the name of Urdu Word Embeddings . Using these embeddings, each word is encoded as a three hundred dimensional vector in the feature space. The next layer is the BiLSTM layer which focuses on the important words in each document and with the help of delete gate; it does not focus on unnecessary words. 150 hidden units are implemented in Bidirectional layer. The output of the BiLSTM layer is then fed as input into the CNN layer. CNN is best known to extract features from the text. A single CNN conv2d layer with three multi size filters i.e. 3, 4 and 5 each of filter size 100 is implemented. Different sized multi filters are used to capture several feature maps of every filter size.

Maxpooling is then applied on each feature map. After this process, the pooled features are concatenated to create a feature vector for dense layer. As a matter of fact, small data sets are more likely to face over fitting problem. Therefore, to overcome the issue of over fitting, L2 regularization is applied. Regularization is intended to manage a complex neural network that avoids overfitting as it impacts the performance of various deep learning models. We have used dropout and L2 regularization. It penalizes large weights so that neural network can be optimized.

Optimization is used for training deep learning models for updating the parameters of neural network across multiple iterations. It helps to improve the performance of model. The opted parameters are Dropout rate, Batch Size, Activation Function and Learning Rate. With the best gained parameters, training process is completed. Details of optimized parameters are shown in Table 3. The training process is done with 32-batch size and 0.5 dropout rate. Softmax is selected as an activation function in the parameter setting.

All the experimentation is conducted in Google Colab’s GPU environment. Through Colab, Google provides 13 GB RAM and 108 GB disk space. For assessing the performance of our proposed model effectively, we focus on identifying the best performing parameters of the model and then make a comparison with the BiLSTM, CNN, CNN-BiLSTM and BERT as baseline deep learning models on two data sets.

\begin{table}  
\centering
\begin{tabular}{ccccc}\toprule
Parameters  &  IMDB \textsubscript{1} & IMDB\textsubscript{2} & IMDB \textsubscript{3} & VTC \\ \midrule
Dropout & 0.6 & 0.8 & 0.8 & 0.5  \\
Batch Size & 32 & 32 & 32 & 32 \\
Activation Function & Softmax & Softmax & Softmax & Softmax  \\
Learning Rate & 2e-05 & 0.001 & 0.0001 & 2e-05\\
\bottomrule

\end{tabular}
\caption{\textbf{Optimized Parameters for BiLSTM-SLMFCNN on VTC and IMDB Data Sets}}
\end{table}

\subsection{Evaluation Metrics:}
Various evaluation metrics are widely used by researchers to evaluate the quality of classification model. In this paper, commonly used performance measures i.e. F1 Score, Accuracy and Area Under Curve(AUC) have been used.

F1 score is known as the weighted average of precision and recall. It takes into consideration both false positives (FP) and false negatives (FN) and it is calculated as follows:
\begin{equation}
F1 - Score = 2 * \frac{Precision * Recall}{Precision + Recall }
\end{equation}

Accuracy of classification model is the fraction of those predictions that model predicts correctly. It is calculated as:
\begin{equation}
Accuracy = \frac{TP +TN}{TP +TN + FP +FN}
\end{equation}

True Positive (TP) refers to model's outcome when it correctly predicts the satisfied class whereas True Negative (TN) refers to model's outcome when it correctly predicts the unsatisfied class. False Positive (FP) refers to model's outcome when it incorrectly predicts the satisfied class whereas False Negative (FN) refers to model's outcome when it incorrectly predicts the unsatisfied class.

Area Under Curve - Receiver Operating Characteristic (AUC-ROC) curve is an evaluation measure that helps us to visualize the performance of binary classification problem. It is used to check the measure of goodness of model in distinguishing two classes. It shows the relationship between sensitivity (True Positive rate) and specificity (False Positive rate) where 
\begin{equation}
Sensitivity = \frac{TP}{TP + FN}
\end{equation}
and 
\begin{equation}
Specificity = \frac{FP}{TN + FP}
\end{equation}

\section{RESULTS AND DISCUSSION:}

\begin{figure*}[!ht]
\centering
\includegraphics[width=1.0\linewidth, height=0.65\linewidth]{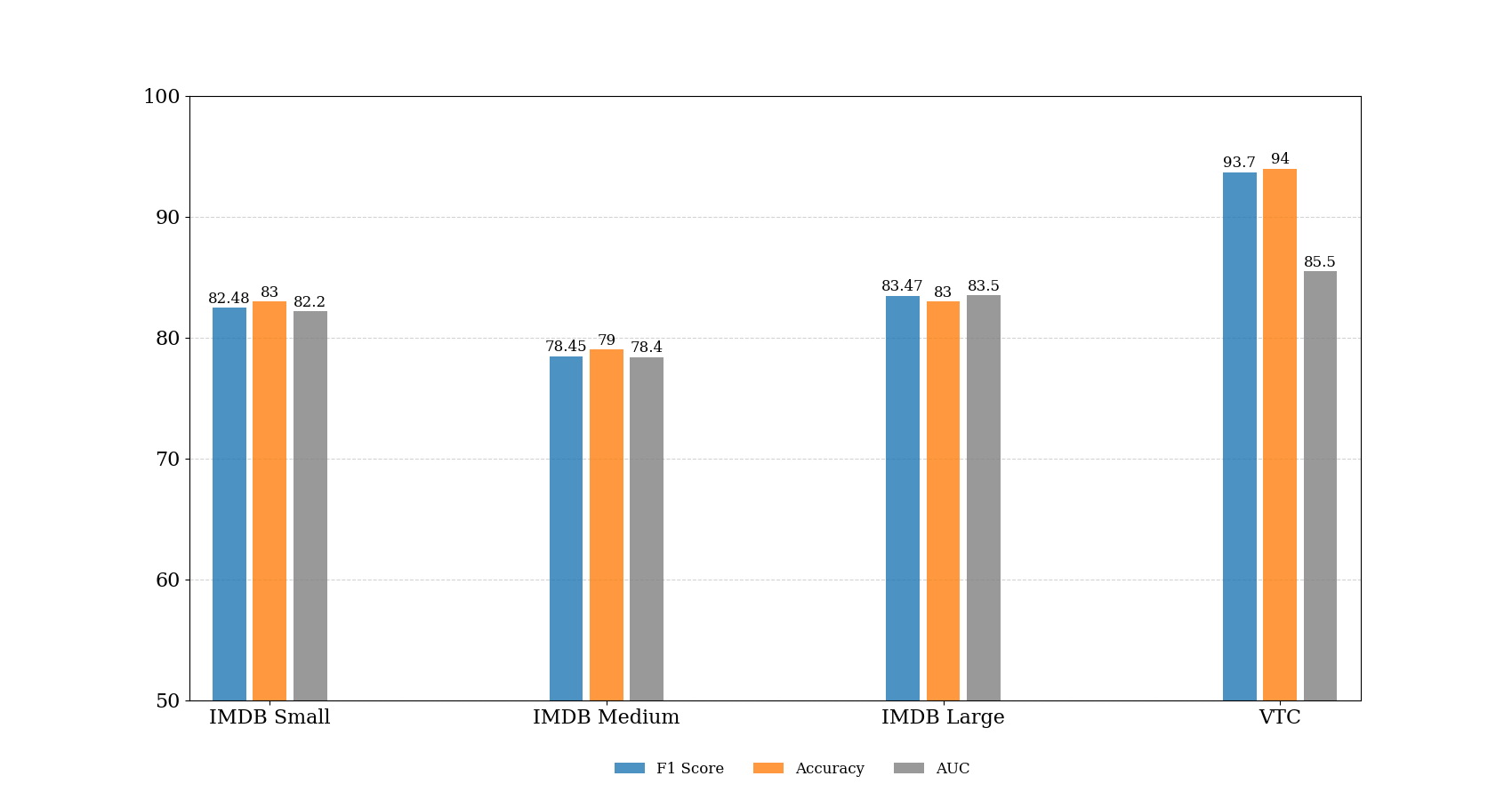}
	\caption{\textbf{Results Showing Performance of BiLSTM-SLMFCNN on VTC Data Set and IMDB Data Sets}}
	\label{fig:Figure 7}
\end{figure*}

During this study, the proposed model and other baseline models were applied on two data sets. In this section, the results of these techniques are discussed in detail.

The proposed model has outperformed other baseline models for all data sets. The VTC data set is small and the core purpose to conduct this research is to check the performance of deep learning models on small Urdu data set. For Urdu sentiment analysis, only machine learning side has been explored till now. As per our knowledge, this is first ever kind of research in Urdu sentiment analysis domain which focuses on the Urdu sentiment classification at document level by applying frequently used deep learning architectures. Results are carefully compared and discussed on the basis of following key factors:
\begin{itemize}
    \item Dataset size
    \item Level of imbalance
    \item Datasets Comparison
    \item Models Comparison
\end{itemize}  

\subsection{Dataset Size: }
The experimental study shows that BiLSTM-SLMFCNN also works well on small sized data set. Large data sets have always been in the consideration of research community; however, it can sometimes be impossible to collect huge data set in a native language. Its high time to test deep learning capabilities on small sized data sets. To break this barrier, we have conducted this experiment so that small data sets may also be considered as important as huge data sets in the deep learning research domain. The results in Table 4 shows that BiLSTM-SLMFCNN has highest F1 score, accuracy and AUC value for all data sets. The comparison of our proposed model and other baseline models is shown in Table 4. 

IMDB small data set has shown comparatively less F1 score and accuracy than VTC data set besides being almost similar in size. However, BiLSTM-SMLFCNN performance on all three IMDB data sets shows that IMDB Large data set outperforms other two data sets in terms of F1 score and accuracy. This apparently indicates that data set size does not affect model’s overall accuracy and F1 metric.

Various sizes of IMDB movie reviews data set have been used to analyze the effect of data size on the accuracy of model. As seen from the results, it is estimated that BiLSTM-SLMFCNN shows best accuracy on IMDB large sized data set. It is due to the reason that large data set has more features and diverse reviews make it possible for the model to learn new words and predict in a better way. IMDB movie review small sized dataset is equivalent in size with VTC’s data set but comparatively shows less accuracy for BiLSTM-SLMFCNN. IMDB movie reviews small and large sized data sets have similar accuracy i.e. 83{\%} and IMDB medium shows 79{\%} accuracy.

According to AUROC curve, higher its value, better is the model at differentiating between two classes. BiLSTM-SLMFCNN gives highest value of AUROC when applied on VTC data set. ROC curve for VTC and IMDB small, medium and large data sets is shown in Figure 5.1. It shows that deep learning models also give satisfied results on small sized data sets in comparison to large sized data sets that have always been considered by researchers. On the basis of results, it is evident that our model works equally well on small, medium and large sized data sets.  

\subsection{Level of Imbalance: }
Our proposed model BILSTM-SLMFCNN gives highest F1 score and accuracy when applied on VTC and variable-sized IMDB data sets. BiLSTM-SLMFCNN has gained F1 score, accuracy and AUC value of 93.7, and 94{\%} and 85.5 respectively when applied on VTC . Although VTC is a slightly imbalanced dataset but our proposed model has the ability to effectively handle imbalance dataset of varying imbalance level and different sizes. In the case of the imbalanced datasets, accuracy is not considered as a good performance measure \cite{chen2018comparison}. For BILSTM-SLMFCNN, the difference between F1 score and accuracy is less for
all datasets as compared to other models. On the basis of comparative analysis, BILSTM-SLMFCNN works well on imbalanced dataset as well.

\subsection{Datasets Comparison: }
The reason that our proposed model works best on VTC data set as compared to IMDB data sets is that VTC data set is very specific to vehicle tracking related user queries while IMDB data set relates to automatically translated movie reviews and it may not have been translated up to desired precision level. As VTC data set has been carefully transcribed and cross-checked, hence it gives better results when different deep learning architectures are applied on it.

\subsection{Models Comparison: }
Other than BiLSTM-SLMFCNN, BiLSTM, CNN CNN-BiLSTM and BERT have also shown good results on both data sets. The ability of CNN to extract text features with help of convolutional layers makes it a powerful deep learning model. Furthermore, BERT also tend to show good performance on small and medium sized data sets as it is a deep bidirectional transformer model that uses Masked Language Model (MLM) for masking words and Next Sentence Prediction (NSP) to learn bidirectional context within a sentence and across different sentences in an efficient manner. However, it has not shown promising results on IMSB Large Urdu data set.

To conclude, the proposed model BiLSTM-SLMFCNN has an ultimate advantage over other deep learning models that the technique used in it is relatively straightforward and automatic in the sense that feature engineering is not involved in the whole process. Hence, it saves implementation cost, time and tends to improve the performance of BiLSTM-SLMFCNN model. The conducted research plays the role of benchmark study in Document level Urdu sentiment analysis using deep learning methods along with the contribution of a proposed hybrid technique.

\begin{table}[!ht]%

\begin{tabular}{|c|c|c|c|c|c|c|c|c|c|c|c|c|}
\hline 

Models/ Datasets &  \multicolumn{3}{c|}{IMDB Small}  &  \multicolumn{3}{c|}{IMDB Medium} &
\multicolumn{3}{c|}{IMDB Large} &
\multicolumn{3
}{c|}{VTC}\\
\hline
\hline
 & F1 & A & AUC & F1 & A & AUC & F1 & A & AUC & F1 & A & AUC \\
\hline
BiLSTM & 67.26 & 51 & 50 & 66.95 & 50 & 50 & 66.95 & 50 & 50 & 71.55 & 80 & 50\\
\hline
CNN & 77.98 & 78 & 78 & 78.22 & 78.2 & 78 & 82.16 & 82 &82.2 & 89.32 & 91 & 76\\    
\hline
CNN-BiLSTM & 64.57 & 67 & 67.7 & 75.46 & 76 & 75.8 & 79.78 & 80 & 79.9 & 88.27 & 90 & 74\\ 
\hline
BERT & 70.48 & 71 & 70 & 62.6 & 63 & 62 & 33.60 & 50 & 50 & 93.53 & 93 & 89\\  
\hline
\textbf{BiLSTM-SLMFCNN} & \textbf{82.48}  & \textbf{83} & \textbf{82.2} & \textbf{78.45} & \textbf{79} & \textbf{78.4} & \textbf{83.47} & \textbf{83} & \textbf{83.5} & \textbf{93.7} & \textbf{94} & \textbf{85.5}
\\
\hline
\end{tabular}
\begin{tabular}{|c|c|c|c|c|}
\hline
\addlinespace[0.15cm]
A - Accuracy &
AUC - Area under curve\\
\hline
\end{tabular}
\caption{\textbf{Results showing performance of BiLSTM-SLMFCNN and other Deep learning models on VTC and IMDB data sets}}
\end{table}



\begin{figure}
 \begin{subfigure}{0.49\textwidth}
     \includegraphics[width=\textwidth]{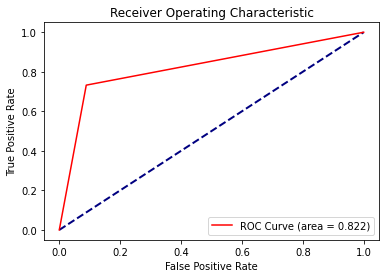}
     \caption{IMDB Small}
     \label{fig:a}
 \end{subfigure}
 \hfill
 \begin{subfigure}{0.49\textwidth}
     \includegraphics[width=\textwidth]{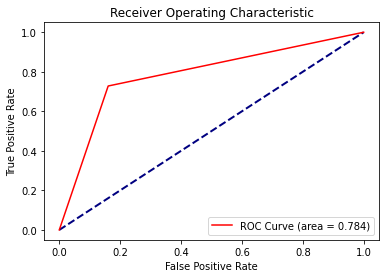}
     \caption{IMDB Medium}
     \label{fig:b}
 \end{subfigure}
 
 \medskip
 \begin{subfigure}{0.49\textwidth}
     \includegraphics[width=\textwidth]{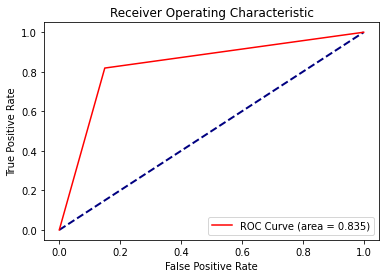}
     \caption{IMDB Large}
     \label{fig:c}
 \end{subfigure}
 \hfill
 \begin{subfigure}{0.49\textwidth}
     \includegraphics[width=\textwidth]{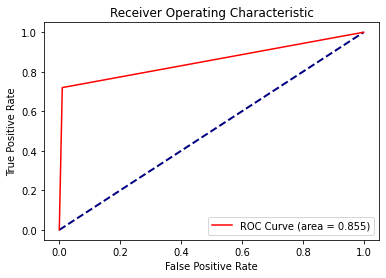}
     \caption{VTC}
     \label{fig:d}
 \end{subfigure}

 \caption{\textbf{ROC Curve for IMDB and VTC Datasets on BiLSTM-SLMFCNN}}
 \label{fig:Figure8}

\end{figure}

The proposed model BiLSTM-SLMFCNN has an ultimate advantage over other deep learning models that the technique used in it is relatively straightforward and automatic in the sense that feature engineering is not involved in the whole process. Hence, it saves implementation cost, time and tends to improve the performance of BiLSTM-SLMFCNN model. We have used average max-pooling instead of simple maxpooling layer which
gave better results in terms of accuracy and F1 score from baseline approaches. Our proposed model used less number of parameters which consumed less memory and is efficient in terms of convolution layers. 

\section{CONCLUSION AND FUTURE WORK:}
In this research, we have made an attempt to gain the attention of research community towards Urdu, a resource-poor language, by applying our proposed neural network based deep leaning model named BiLSTM-SLMFCNN on two Urdu data sets. The conducted research plays the role of benchmark study in document level Urdu SA using DL methods along with the contribution of a proposed hybrid technique.  

Document level Urdu SA is a benchmark study in the Urdu NLP domain. We have also applied four other deep learning models i.e. BiLSTM, CNN, CNN-BiLSTM and BERT on both data sets to compare the performance of our model. We have presented the experimental study that has been opted for generating the results on two data sets i.e. VTC customer support and IMDB movie review data set. When applied on both data sets, the results show that our proposed technique outperformed other baseline models. Irrespective of data size in three different size versions of IMDB data set, our model has gained highest accuracy and F1 score. For finding our parameters, grid search approach has been adopted. The process of finding the optimal parameters is time consuming and exhaustive but it helps to improve the performance of classification model. We have also demonstrated the significance of pre-trained neural Urdu word embedding for document level SA. The proposed technique can also be used for numerous other Urdu text classification tasks such as detecting fake news, abuse detection on social media etc. As Urdu faces the issue of resource scarcity, so there lies wide scope of further research. 

For future work, we intend to integrate lexicon with pretrained neural word embeddings in BiLSTM-SLMFCNN to analyze the effect of using lexicons with DL methods. Comparison of techniques discussed in this paper can also be made on other balanced data sets. Furthermore, incorporating attention based mechanism in order to further improve model’s accuracy on different sized data sets is also a worth exploring research dimension. 

\printcredits


\bibliography{highlights}

\begin{thebibliography}{36}
\expandafter\ifx\csname natexlab\endcsname\relax\def\natexlab#1{#1}\fi

\bibitem[{Abdul-Mageed and Korayem(2010)}]{abdul2010automatic}
Muhammad Abdul-Mageed and Mohammed Korayem. 2010.
\newblock Automatic identification of subjectivity in morphologically rich languages: the case of arabic.
\newblock \emph{Computational approaches to subjectivity and sentiment analysis}, 2:2--6.

\bibitem[{Ahmed et~al.(2016)Ahmed, Ali, Khalid, and Kamran}]{ahmed2016framework}
Kashif Ahmed, Mubashir Ali, Shehzad Khalid, and Muhammad Kamran. 2016.
\newblock Framework for urdu news headlines classification.
\newblock \emph{Journal of Applied Computer Science \& Mathematics}, (21).

\bibitem[{Akhter et~al.(2022)Akhter, Jiangbin, Naqvi, Abdelmajeed, and Fayyaz}]{akhter2022exploring}
Muhammad~Pervez Akhter, Zheng Jiangbin, Irfan~Raza Naqvi, Mohammed Abdelmajeed, and Muhammad Fayyaz. 2022.
\newblock Exploring deep learning approaches for urdu text classification in product manufacturing.
\newblock \emph{Enterprise Information Systems}, 16(2):223--248.

\bibitem[{Akhter et~al.(2020)Akhter, Jiangbin, Naqvi, Abdelmajeed, Mehmood, and Sadiq}]{akhter2020document}
Muhammad~Pervez Akhter, Zheng Jiangbin, Irfan~Raza Naqvi, Mohammed Abdelmajeed, Atif Mehmood, and Muhammad~Tariq Sadiq. 2020.
\newblock Document-level text classification using single-layer multisize filters convolutional neural network.
\newblock \emph{IEEE Access}, 8:42689--42707.

\bibitem[{Ali and Ijaz(2009)}]{ali2009urdu}
Abbas~Raza Ali and Maliha Ijaz. 2009.
\newblock Urdu text classification.
\newblock In \emph{Proceedings of the 7th international conference on frontiers of information technology}, pages 1--7.

\bibitem[{Ali et~al.(2016)Ali, Noor, Javed, Aslam, Khan et~al.}]{ali2016salience}
S~Abbas Ali, M~Daniyal Noor, Munir~Ahmed Javed, M~Mohsin Aslam, Omer~Ahmed Khan, et~al. 2016.
\newblock Salience analysis of news corpus using heuristic approach in urdu language.
\newblock \emph{International Journal of Computer Science and Network Security (IJCSNS)}, 16(4):28.

\bibitem[{Amjad et~al.(2017)Amjad, Ishtiaq, Firdous, and Mehmood}]{amjad2017exploring}
Kamran Amjad, Maria Ishtiaq, Samar Firdous, and Muhammad~Amir Mehmood. 2017.
\newblock Exploring twitter news biases using urdu-based sentiment lexicon.
\newblock In \emph{2017 International Conference on Open Source Systems \& Technologies (ICOSST)}, pages 48--53. IEEE.

\bibitem[{Asghar et~al.(2017)Asghar, Khan, Khan, Ahmad, and Khan}]{asghar2017cogemo}
Muhammad~Zubair Asghar, Aurangzeb Khan, Khairullh Khan, Hussain Ahmad, and Imran~Ali Khan. 2017.
\newblock Cogemo: Cognitive-based emotion detection from patient generated health reviews.
\newblock \emph{Journal of Medical Imaging and Health Informatics}, 7(6):1436--1444.

\bibitem[{Awais and Shoaib(2019)}]{awais2019role}
Dr~Muhammad Awais and Dr~Muhammad Shoaib. 2019.
\newblock Role of discourse information in urdu sentiment classification: A rule-based method and machine-learning technique.
\newblock \emph{ACM Transactions on Asian and Low-Resource Language Information Processing (TALLIP)}, 18(4):1--37.

\bibitem[{Bibi et~al.(2019)Bibi, Qamar, Ansar, and Shaheen}]{bibi2019sentiment}
Raheela Bibi, Usman Qamar, Munazza Ansar, and Asma Shaheen. 2019.
\newblock Sentiment analysis for urdu news tweets using decision tree.
\newblock In \emph{2019 IEEE 17th international conference on software engineering research, management and applications (SERA)}, pages 66--70. IEEE.

\bibitem[{Bloom and Argamon(2010)}]{bloom2010automated}
Kenneth Bloom and Shlomo Argamon. 2010.
\newblock Automated learning of appraisal extraction patterns.
\newblock In \emph{Corpus-linguistic applications}, pages 249--260. Brill Rodopi.

\bibitem[{Chen et~al.(2018)Chen, McKeever, and Delany}]{chen2018comparison}
Hao Chen, Susan McKeever, and Sarah~Jane Delany. 2018.
\newblock A comparison of classical versus deep learning techniques for abusive content detection on social media sites.
\newblock In \emph{Social Informatics: 10th International Conference, SocInfo 2018, St. Petersburg, Russia, September 25-28, 2018, Proceedings, Part I 10}, pages 117--133. Springer.

\bibitem[{Daud et~al.(2017)Daud, Khan, and Che}]{daud2017urdu}
Ali Daud, Wahab Khan, and Dunren Che. 2017.
\newblock Urdu language processing: a survey.
\newblock \emph{Artificial Intelligence Review}, 47:279--311.

\bibitem[{Gao et~al.(2019)Gao, Feng, Song, and Wu}]{gao2019target}
Zhengjie Gao, Ao~Feng, Xinyu Song, and Xi~Wu. 2019.
\newblock Target-dependent sentiment classification with bert.
\newblock \emph{Ieee Access}, 7:154290--154299.

\bibitem[{Ghulam et~al.(2019)Ghulam, Zeng, Li, and Xiao}]{ghulam2019deep}
Hussain Ghulam, Feng Zeng, Wenjia Li, and Yutong Xiao. 2019.
\newblock Deep learning-based sentiment analysis for roman urdu text.
\newblock \emph{Procedia computer science}, 147:131--135.

\bibitem[{Goularas and Kamis(2019)}]{goularas2019evaluation}
Dionysis Goularas and Sani Kamis. 2019.
\newblock Evaluation of deep learning techniques in sentiment analysis from twitter data.
\newblock In \emph{2019 International Conference on Deep Learning and Machine Learning in Emerging Applications (Deep-ML)}, pages 12--17. IEEE.

\bibitem[{Haider(2018)}]{haider2018urdu}
Samar Haider. 2018.
\newblock Urdu word embeddings.
\newblock In \emph{Proceedings of the Eleventh International Conference on Language Resources and Evaluation (LREC 2018)}.

\bibitem[{Hassan and Shoaib(2018)}]{hassan2018opinion}
Muhammad Hassan and Muhammad Shoaib. 2018.
\newblock Opinion within opinion: segmentation approach for urdu sentiment analysis.
\newblock \emph{Int. Arab J. Inf. Technol.}, 15(1):21--28.

\bibitem[{Hochreiter(1997)}]{hochreiter1997long}
S~Hochreiter. 1997.
\newblock Long short-term memory.
\newblock \emph{Neural Computation MIT-Press}.

\bibitem[{Khan et~al.(2019)Khan, Jan, and Farman}]{khan2019deep}
Murad Khan, Bilal Jan, and Haleem Farman. 2019.
\newblock \emph{Deep learning: convergence to big data analytics}.
\newblock Springer.

\bibitem[{Khattak et~al.(2021)Khattak, Asghar, Saeed, Hameed, Hassan, and Ahmad}]{khattak2021survey}
Asad Khattak, Muhammad~Zubair Asghar, Anam Saeed, Ibrahim~A Hameed, Syed~Asif Hassan, and Shakeel Ahmad. 2021.
\newblock A survey on sentiment analysis in urdu: A resource-poor language.
\newblock \emph{Egyptian Informatics Journal}, 22(1):53--74.

\bibitem[{Liu(2012)}]{liu2012sentiment}
Bing Liu. 2012.
\newblock Sentiment analysis and opinion mining.
\newblock \emph{Synthesis lectures on human language technologies}, 5(1):1--167.

\bibitem[{Mathew et~al.(2021)Mathew, Amudha, and Sivakumari}]{mathew2021deep}
Amitha Mathew, P~Amudha, and S~Sivakumari. 2021.
\newblock Deep learning techniques: an overview.
\newblock \emph{Advanced Machine Learning Technologies and Applications: Proceedings of AMLTA 2020}, pages 599--608.

\bibitem[{Melville et~al.(2009)Melville, Gryc, and Lawrence}]{melville2009sentiment}
Prem Melville, Wojciech Gryc, and Richard~D Lawrence. 2009.
\newblock Sentiment analysis of blogs by combining lexical knowledge with text classification.
\newblock In \emph{Proceedings of the 15th ACM SIGKDD international conference on Knowledge discovery and data mining}, pages 1275--1284.

\bibitem[{Mukhtar et~al.(2017)Mukhtar, Khan, and Chiragh}]{mukhtar2017effective}
Neelam Mukhtar, Mohammad~Abid Khan, and Nadia Chiragh. 2017.
\newblock Effective use of evaluation measures for the validation of best classifier in urdu sentiment analysis.
\newblock \emph{Cognitive Computation}, 9:446--456.

\bibitem[{Mukhtar et~al.(2018)Mukhtar, Khan, Chiragh, and Nazir}]{mukhtar2018identification}
Neelam Mukhtar, Mohammad~Abid Khan, Nadia Chiragh, and Shah Nazir. 2018.
\newblock Identification and handling of intensifiers for enhancing accuracy of urdu sentiment analysis.
\newblock \emph{Expert Systems}, 35(6):e12317.

\bibitem[{Nassr et~al.(2019)Nassr, Sael, and Benabbou}]{nassr2019comparative}
Zineb Nassr, Nawal Sael, and Faouzia Benabbou. 2019.
\newblock A comparative study of sentiment analysis approaches.
\newblock In \emph{Proceedings of the 4th International Conference on Smart City Applications}, pages 1--8.

\bibitem[{Rakhlin(2016)}]{rakhlin2016convolutional}
A~Rakhlin. 2016.
\newblock Convolutional neural networks for sentence classification.
\newblock \emph{GitHub}, 6:25.

\bibitem[{Rhanoui et~al.(2019)Rhanoui, Mikram, Yousfi, and Barzali}]{rhanoui2019cnn}
Maryem Rhanoui, Mounia Mikram, Siham Yousfi, and Soukaina Barzali. 2019.
\newblock A cnn-bilstm model for document-level sentiment analysis.
\newblock \emph{Machine Learning and Knowledge Extraction}, 1(3):832--847.

\bibitem[{Riaz(2012)}]{riaz2012comparison}
Kashif Riaz. 2012.
\newblock Comparison of hindi and urdu in computational context.
\newblock \emph{Int J Comput Linguist Nat Lang Process}, 1(3):92--97.

\bibitem[{Schuster and Paliwal(1997)}]{schuster1997bidirectional}
Mike Schuster and Kuldip~K Paliwal. 1997.
\newblock Bidirectional recurrent neural networks.
\newblock \emph{IEEE transactions on Signal Processing}, 45(11):2673--2681.

\bibitem[{Syed et~al.(2011{\natexlab{a}})Syed, Aslam, and Martinez-Enriquez}]{syed2011adjectival}
Afraz~Z Syed, M~Aslam, and A~Martinez-Enriquez. 2011{\natexlab{a}}.
\newblock Adjectival phrases as the sentiment carriers in the urdu text.
\newblock \emph{Journal of American Science}, 7(3):644--652.

\bibitem[{Syed et~al.(2010)Syed, Aslam, and Martinez-Enriquez}]{syed2010lexicon}
Afraz~Z Syed, Muhammad Aslam, and Ana~Maria Martinez-Enriquez. 2010.
\newblock Lexicon based sentiment analysis of urdu text using sentiunits.
\newblock In \emph{Advances in Artificial Intelligence: 9th Mexican International Conference on Artificial Intelligence, MICAI 2010, Pachuca, Mexico, November 8-13, 2010, Proceedings, Part I 9}, pages 32--43. Springer Berlin Heidelberg.

\bibitem[{Syed et~al.(2011{\natexlab{b}})Syed, Aslam, and Martinez-Enriquez}]{syed2011sentiment}
Afraz~Zahra Syed, Muhammad Aslam, and Ana~Maria Martinez-Enriquez. 2011{\natexlab{b}}.
\newblock Sentiment analysis of urdu language: handling phrase-level negation.
\newblock In \emph{Advances in Artificial Intelligence: 10th Mexican International Conference on Artificial Intelligence, MICAI 2011, Puebla, Mexico, November 26-December 4, 2011, Proceedings, Part I 10}, pages 382--393. Springer.

\bibitem[{Vaswani(2017)}]{vaswani2017attention}
A~Vaswani. 2017.
\newblock Attention is all you need.
\newblock \emph{Advances in Neural Information Processing Systems}.

\bibitem[{Zia et~al.(2015)Zia, Akhter, and Abbas}]{zia2015comparative}
Tehseen Zia, Muhammad~Pervez Akhter, and Qaiser Abbas. 2015.
\newblock Comparative study of feature selection approaches for urdu text categorization.
\newblock \emph{Malaysian Journal of Computer Science}, 28(2):93--109.

\end{thebibliography}
\bibliographystyle{model1-num-names}

\end{document}